\documentclass{article}
\pdfoutput=1
\usepackage{microtype}
\usepackage{graphicx}
\usepackage{subcaption}
\usepackage{caption} 
\usepackage{booktabs} 
\usepackage{multirow} 
\usepackage[table]{xcolor}

\usepackage[inline]{enumitem}

\usepackage{natbib}
\setcitestyle{authoryear,round,comma,aysep={,},yysep={;}}

\usepackage{hyperref}

\usepackage[preprint]{icml2026}

\usepackage{amsmath}
\usepackage{amssymb}
\usepackage{mathtools}
\usepackage{amsthm}

\usepackage[capitalize,noabbrev]{cleveref}

\theoremstyle{plain}

\theoremstyle{definition}

\theoremstyle{remark}

\usepackage[textsize=tiny]{todonotes}

\icmltitlerunning{Submission and Formatting Instructions for ICML 2026}

\begin{document}

\twocolumn[
  \icmltitle{Geo-Code: A Code Framework for Reverse Code Generation from Geometric Images Based on Two-Stage Multi-Agent Evolution}



  \icmlsetsymbol{equal}{*}

 \begin{icmlauthorlist}
    \icmlauthor{Zhenyu Wu}{bnu}
    \icmlauthor{yanxi long}{bnu}
    \icmlauthor{Jian Li}{bnu}
    \icmlauthor{Hua Huang}{bnu}
  \end{icmlauthorlist}

  \icmlaffiliation{bnu}{School of Artificial Intelligence, Beijing Normal University, Beijing, China}

  \icmlcorrespondingauthor{Jian Li}{jli@bnu.edu.cn}

  \icmlkeywords{Machine Learning, ICML, Code Generation, Geometric Reasoning}

  \vskip 0.3in
]



\printAffiliationsAndNotice{}  


\section{Abstract}
Program code serves as a bridge linking vision and logic, providing a feasible supervisory approach for enhancing the multimodal reasoning capability of large models through geometric operations (e.g., auxiliary line construction and perspective transformation). Nevertheless, current inverse graphics methods face tremendous challenges in accurately reconstructing complex geometric details, which often results in the loss of key geometric constraints or structural distortion. To address this bottleneck, we propose Geo-coder—the first inverse programming framework for geometric images based on a multi-agent system. Our method innovatively decouples the process into geometric modeling via pixel-wise anchoring and metric-driven code evolution: Stage 1 leverages the complementary advantages of visual operators and large models to achieve precise capture of pixel coordinates and visual attributes; Stage 2 introduces a synthesis-rendering-validation closed loop, where bidirectional visual feedback drives the self-correction of code. Extensive experiments demonstrate that Geo-coder achieves a substantial lead in both geometric reconstruction accuracy and visual consistency. Notably, by effectively preserving the core geometric semantics, the images reconstructed with our method exhibit equivalent performance to the original ones in multimodal reasoning tasks, which fully validates the robustness of the framework. Finally, to further reduce research costs, we have open-sourced the Geo-coder dataset constructed on the GeoCode framework, which contains more than 1,500 samples. On this basis, we have also open-sourced the GeocodeLM model, laying a solid data and model foundation for subsequent research in this field.
\section{Introduction}

\begin{figure}[htbp]
    \centering  
    \begin{subfigure}{0.18\textwidth}  
        \centering  
        \includegraphics[width=\linewidth]{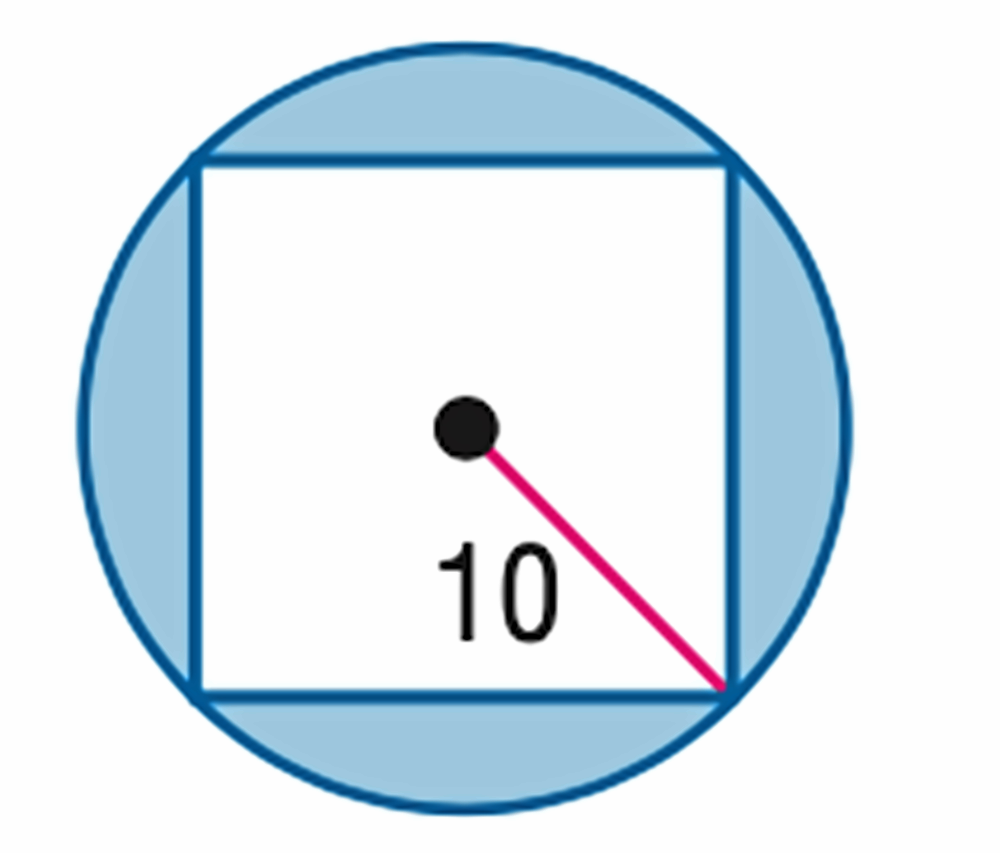}  
        \subcaption{origin}  
        \label{origin}  
    \end{subfigure}
    \hfill  
    \begin{subfigure}{0.18\textwidth}
        \centering
        \includegraphics[width=\linewidth]{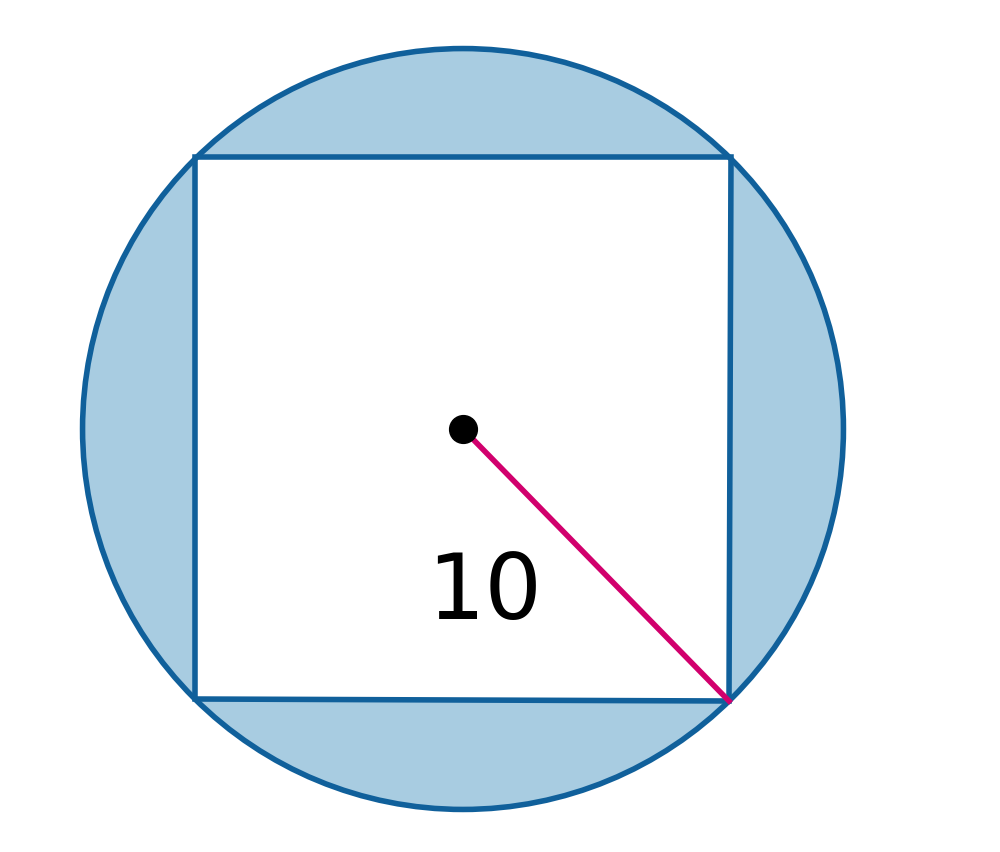}  
        \subcaption{GeoCode}
        \label{Geocode}
    \end{subfigure}
    
    \vspace{0.3cm}  
    \begin{subfigure}{0.18\textwidth}
        \centering
        \includegraphics[width=\linewidth]{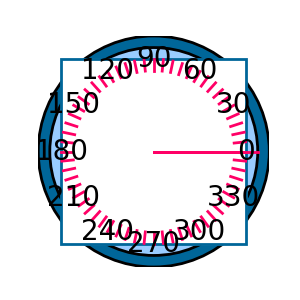}  
        \subcaption{MatPlotCode}
        \label{math}
    \end{subfigure}
    \hfill
    \begin{subfigure}{0.18\textwidth}
        \centering
        \includegraphics[width=\linewidth]{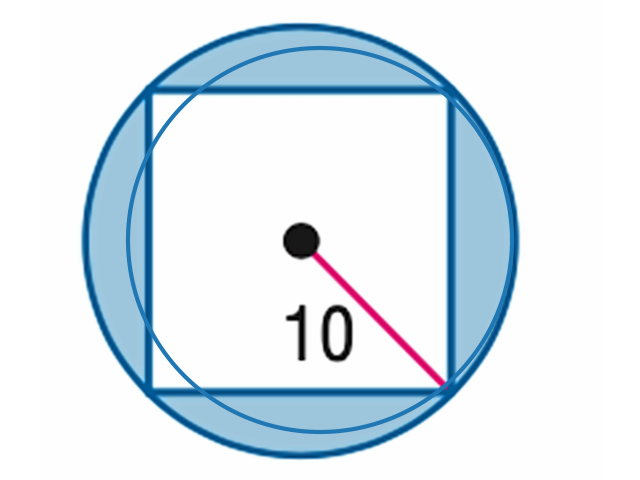}  
        \label{Vthink}
        \subcaption{V-Thinker }
    \end{subfigure}

    \caption{Comparison plot of different reconstruction methods. 
    For more examples, refer to the Appendix. }
    \label{four_figs}  
\end{figure}

In recent years, Multimodal Language Models (MLLMs) have undergone a significant paradigm shift in mathematical reasoning tasks, evolving from pure text generation to \textit{interleaved text-visual reasoning} \cite{hu2024visualsketchpadsketchingvisual,qiao2025vthinkerinteractivethinkingimages,wang2025mathcodervlbridgingvisioncode}. In this process, the enhancement of reasoning capabilities essentially relies on visual information completion through \textit{dynamic geometric manipulation}, such as constructing auxiliary lines or performing affine transformations. Consequently, executable program code has emerged as a \textit{pivotal intermediate modality}, bridging the gap between abstract logical reasoning and precise visual rendering.

However, a critical bottleneck impedes current progress: the vast majority of multimodal datasets lack corresponding executable code representations. Existing inverse graphics methods, which rely on end-to-end visual generation, suffer from two fundamental limitations when reverse-engineering code:

\begin{enumerate}
    \item \textbf{Insufficient Geometric Structural Fidelity}: Constrained by inherent visual perception limits, these methods often treat coordinate prediction as a direct regression task, which is prone to \textit{visual hallucinations}. As illustrated in Figure~\ref{math}, such approaches often introduce extraneous artifacts (e.g., redundant labels) or topological inconsistencies, which fundamentally alter geometric semantics and impair the reasoning accuracy of large models.
    \item \textbf{Loss of Primitive-level Details}: Existing methods struggle to capture \textit{primitive-level nuances}, such as sub-pixel coordinates and precise line thickness. This results in a significant \textit{domain gap} in visual fidelity between the generated images and the original distribution. Although macroscopically similar, these discrepancies fail to preserve the diverse data characteristics inherent in the original datasets, as illustrated in Figure~\ref{fig:detailed_accuracy_single_row}.
\end{enumerate}

To address these challenges, we propose \textbf{Geo-code}, the first Multi-Agent System (MAS) framework designed for geometric image reverse engineering. Unlike conventional approaches, Geo-code innovatively \textit{decouples} the reverse-engineering process into two strategic stages, incorporating novel mechanisms to bridge the neuro-symbolic gap:
\begin{enumerate*}[label=(\arabic*)]
    \item \textbf{Meta-information Extraction Stage via Pixel-wise Anchoring}: 
    Instead of implicitly guessing coordinates, we introduce \textbf{Pixel-wise Anchoring Operators}. By synergizing classical computer vision operators (to extract high-frequency gradient changes like junctions) with the semantic understanding of MLLMs, we establish explicit \textit{visual anchors}. This mechanism solves the symbol grounding problem by binding geometric primitives to concrete pixel data, constructing a precise initial geometric skeleton.
    
    \item \textbf{Program Evolution Stage via Visual Error Projection}: 
    This stage operates as a Synthesis-Rendering-Verification closed loop. Crucially, we propose a \textbf{Visual Error Projection (VEP)} mechanism. Rather than relying on abstract numerical losses, VEP projects the Chamfer Distance between the rendered canvas and the ground truth into interpretable \textit{Visual Difference Maps}. This allows the Refiner Agent to intuitively "see" topological discrepancies and perform gradient-free optimization, iteratively correcting deviations until high-precision alignment is achieved.
\end{enumerate*}

Extensive experiments demonstrate that Geo-code establishes a new \textit{state-of-the-art (SOTA)} on both \textit{pixel-level metrics} (e.g., Hausdorff and Chamfer distances) and \textit{perceptual-level metrics}. Furthermore, we evaluated the efficacy of our reconstructions in downstream multimodal reasoning tasks. The results reveal \textit{zero performance loss} on the GeoSketch, GeoQA, and AuxSolidMath benchmarks, and even a 4\% accuracy gain on the MathVerse dataset, validating the \textit{high robustness} and semantic consistency of our method.

Finally, to facilitate future research, we introduce the \textbf{Geo-code Dataset}, a high-quality benchmark of over 1,500 image-code pairs. Distinct from noisy web-crawled data, every entry has undergone a rigorous \textbf{three-tier verification process}. Leveraging this golden standard data, we fine-tuned the \textbf{GeoCodeLM}, which significantly outperforms baselines in both code generation accuracy and visual fidelity.

The primary contributions of this paper are summarized as follows:
\begin{itemize}
    \item We propose Geo-code, a neuro-symbolic framework that addresses the missing code problem. We introduce two core mechanisms: \textbf{Pixel-wise Anchoring Operators} for precise coordinate grounding and \textbf{Visual Error Projection} for intuition-based iterative refinement.
    \item We demonstrate that our framework achieves SOTA performance in geometric fidelity. Crucially, the reconstructed images maintain \textit{lossless performance} in downstream multimodal reasoning tasks, proving that \textbf{logical rigor} (code) is superior to pure visual alignment.
    \item We release the \textbf{Geo-code Dataset}, a rigorously verified image-code benchmark, and the fine-tuned \textbf{GeoCodeLM}, providing a solid foundation for the community to explore code-based visual reasoning.
\end{itemize}

\section{Related Works}
\label{sec:related_work}

\subsection{Geometric Image Editing}
Recent advancements have explored integrating visual reasoning into the inference process, primarily by editing geometric images—such as adding auxiliary lines—to complete geometric information. Existing methods can be categorized into three primary approaches. 
First, \textbf{code-based methods} have gained significant attention. \textit{Visual Sketchpad} \cite{hu2024visualsketchpadsketchingvisual} pioneered the use of agents that employ code as an intermediate modality for dynamic image editing, empowering models with fundamental drawing capabilities to reveal latent geometric structures. Subsequent works, such as \textit{V-Thinker} \cite{qiao2025vthinkerinteractivethinkingimages}, \textit{DeepSketcher} \cite{zhang2025deepsketcherinternalizingvisualmanipulation}, \textit{CodePlot-CoT} \cite{duan2025codeplotcotmathematicalvisualreasoning}, \textit{GeometryZero} \cite{wang2025geometryzeroimprovinggeometrysolving}, \textit{MathCanvas} \cite{shi2025mathcanvasintrinsicvisualchainofthought}, and \textit{MathCoder-VL} \cite{wang2025mathcodervlbridgingvisioncode}, utilize fine-tuning to enable models to perform reasoning through image editing. However, these methods rely heavily on the availability of pre-existing image code. When the original images lack corresponding code representations, direct editing becomes computationally intractable or impossible.
Second, \textbf{pixel-based methods} utilize image-to-image generative models for direct editing; nevertheless, they often struggle to comprehend deep geometric structures, frequently leading to \textit{geometric hallucinations}. 
Finally, \textbf{text-based and neuro-symbolic methods} offer alternative paradigms. For instance, \textit{GeoVLMath} \cite{guo2026geovlmathenhancinggeometryreasoning} avoids direct image editing by taking both the original image and textual descriptions of auxiliary lines as input, while \textit{Geoint-R1} \cite{wei2025geointr1formalizingmultimodalgeometric} implements auxiliary line reasoning through specialized learning methods. 
Our research focuses on a critical prerequisite for these code-based methods: the accurate generation of corresponding code from an input image.

\subsection{Image-to-Code Construction}
The transformation of images into executable code is a foundational task for visual reasoning. Current methodologies encompass three main strategies, yet each faces significant limitations.
One prominent direction is \textbf{data-driven supervised fine-tuning (SFT)}, where works like \textit{CodePlot-CoT} \cite{duan2025codeplotcotmathematicalvisualreasoning} and \textit{MathCoder-VL} \cite{wang2025mathcodervlbridgingvisioncode} construct large-scale image-to-code datasets to train specialized models. However, the efficacy of these methods is bottlenecked by data quality: existing datasets typically provide only final code outputs, lacking explicit intermediate geometric meta-information (e.g., primitive constraints and attributes). Furthermore, the inherent lack of precision in the training data often leads models to learn superficial code patterns rather than precise geometric construction logic.
Another approach involves \textbf{modular and pipeline frameworks}. For example, \textit{GeoSketch} \cite{weng2025geosketchneuralsymbolicapproachgeometric} uses a YOLO-based detector to convert geometric information into neuro-symbols mapped onto fixed templates, while \textit{V-Thinker} \cite{qiao2025vthinkerinteractivethinkingimages} employs a similar intermediate pipeline step. These methods often rely on rigid templates or unidirectional pipelines, suffering from the absence of a closed-loop verification mechanism. Consequently, perception errors from early stages propagate directly to the final code without opportunity for correction.
Lastly, \textbf{zero-shot generation} attempts to leverage the powerful visual capabilities of general Multimodal Large Language Models (MLLMs). Despite their flexibility, MLLMs face inherent perception limits regarding sub-pixel coordinates, frequently resulting in geometric hallucinations.
To address these gaps, this paper introduces a systematic framework that synergizes the precision of classical operators with the semantic understanding of MLLMs, reinforced by a rigorous verification loop for high-fidelity reconstruction.

\section{Methodology}

\begin{figure*}
    \centering
    \includegraphics[width=1\linewidth]{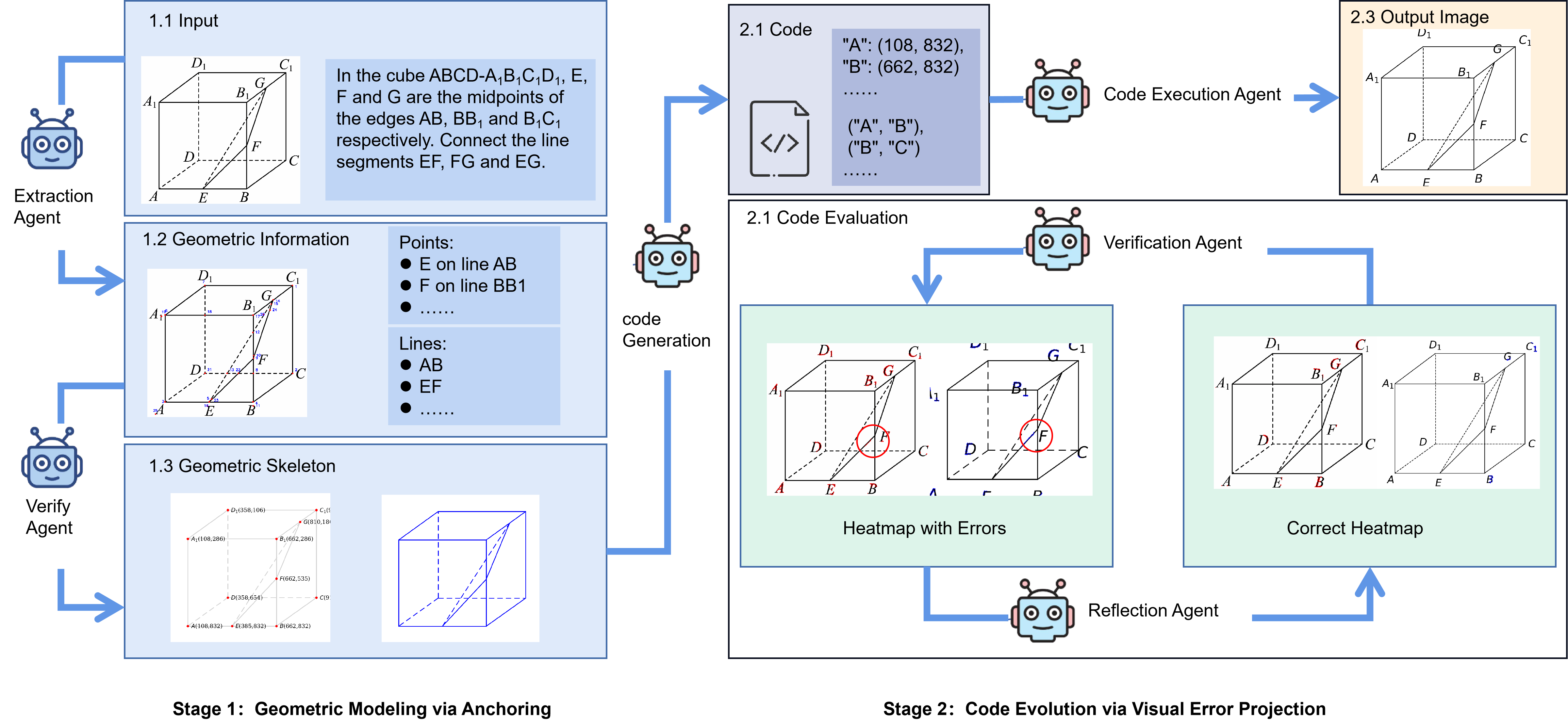}
    \caption{The Geo-code framework operates in two strategic phases. Phase 1 (Geometric Modeling via Anchoring) employs the Geometric Extraction Agent (1.2) to derive geometric attributes via \textbf{Pixel-wise Anchoring}, followed by the Visual Verification Agent (1.3) to synthesize the skeleton. Phase 2 translates this skeleton into executable code, utilizing the Hybrid Inspection and Reflective Correction Agents (2.2) to iteratively refine the output using the \textbf{Code Evolution via Visual Error Projection} mechanism for precise alignment.}
    \label{Main_img}
\end{figure*}

\subsection{Problem Definition}

In this paper, we formulate the geometric image reconstruction task as a \textbf{Neuro-Symbolic Inverse Program Synthesis} problem. Given the visual representation of an original image and its corresponding text description, the objective is to automatically deduce executable geometric program code such that its rendered result recovers the input intention in terms of geometric structure, mathematical logic, and visual fidelity.

The system input is defined as \textbf{multi-modal perceptual data}, denoted as $X=\{\mathcal{I}_{obs}, \mathcal{T}\}$, where:
\begin{itemize}
    \item $\mathcal{I}_{obs}$ represents the \textbf{Observed Image}, i.e., the rasterized visual observation to be reconstructed;
    \item $\mathcal{T}$ represents \textbf{Semantic Constraints}, used to resolve geometric ambiguities (e.g., textual prompts specifying right triangle or tangent circles).
\end{itemize}

The expected output is an \textbf{executable geometric instance}, denoted as $Y=\{\mathcal{C}, \mathcal{I}_{rec}\}$. This output consists of two components: the program code module $\mathcal{C}$ and the reconstructed graphics module $\mathcal{I}_{rec}$. Specifically, $\mathcal{C}$ comprises a geometric parameter set $\Theta$ (containing grounded keypoint coordinates and topology) and a logical control flow $\mathcal{L}$ (program structure). $\mathcal{I}_{rec}$ is the final visualization generated by executing code $\mathcal{C}$ via a deterministic renderer.

The entire geometric reconstruction system aims to learn the inverse mapping relationship from perceptual data $X$ to the program instance $Y$, formally expressed as:
\begin{equation}
    (\mathcal{C}, \mathcal{I}_{rec}) = \Phi(X; \Omega)
\end{equation}
where $\Omega$ represents the system parameters. To achieve the translation from raster pixels to vector programs, the core objective is to maximize quality across three dimensions: \textbf{Geometric Fidelity}, \textbf{Mathematical Self-consistency}, and \textbf{Semantic Consistency}.

\subsection{Methodology Framework}

To solve the mapping $\Phi$, we propose a \textbf{Multi-Agent Staged Evolution Framework}. This framework decouples the complex inference process into two serial phases, utilizing the interaction of six specialized agents to achieve closed-loop optimization from \textit{explicit visual grounding} to \textit{iterative code refinement}.

\subsubsection{Phase I: Multi-modal Geometric Modeling via Anchoring}

This phase is coordinated by two perception-oriented agents, aiming to extract high-precision geometric priors from the unstructured image $\mathcal{I}_{obs}$ and text $\mathcal{T}$ to construct the Geometric Skeleton $\mathcal{S}_{geo}$.

\paragraph{(1) Geometric Extraction Agent ($\pi_{ext}$)}
This agent performs Coarse Reading by implementing our proposed \textbf{Pixel-wise Anchoring Mechanism}. It synergizes classical computer vision operators with large model reasoning to parallelly extract topological logic and physical features.
\begin{itemize}
    \item \textbf{Topology Discovery:} It maximizes the mutual information between logical constraints $\mathcal{R}$ and input data to identify relationships such as parallelism and orthogonality:
    \begin{equation}
        \mathcal{R}^* = \pi_{ext}^{logic}(\mathcal{I}_{obs}, \mathcal{T}) = \underset{\mathcal{R}}{\arg\max} \, I(\mathcal{R}; \mathcal{I}_{obs}, \mathcal{T})
    \end{equation}
    \item \textbf{Pixel-wise Anchoring:} Unlike direct regression, it utilizes specific image gradient operators (e.g., junction detection) to extract potential geometric feature points as physical anchors:
    \begin{equation}
        \mathbf{P}_{raw} = \pi_{ext}^{vision}(\mathcal{I}_{obs})
    \end{equation}
\end{itemize}

\paragraph{(2) Visual Verification Agent ($\pi_{ver}$)}
This agent acts as a Semantic Filter to solve the symbol grounding problem. It maps $\mathbf{P}_{raw}$ back to the original image context, performing denoising and validation.
\begin{equation}
\begin{aligned}
    \mathbf{P}^* &= \pi_{ver}(\mathbf{P}_{raw}, \mathcal{I}_{obs}) \\
    &= \{ p \in \mathbf{P}_{raw} \mid \text{Verify}(p, \mathcal{I}_{obs}) = \text{True} \}
\end{aligned}
\end{equation}

Phase I ultimately outputs the structured representation $\mathcal{S}_{geo} = \{ \mathbf{P}^*, \mathcal{R}^* \}$, which serves as the grounded input for Phase II.

\subsubsection{Phase II: Code Evolution via Visual Error Projection}

This phase constitutes a closed-loop system driven by four execution and optimization agents. It synthesizes and refines the code through a \textbf{Generate-Execute-Inspect-Correct} iterative mechanism, leveraging visual feedback to guide optimization.

\paragraph{(1) Code Generation Agent ($\pi_{gen}$)}
Serving as the system's Initializer, this agent constructs the initial code $\mathcal{C}^{(0)}$ based on the grounded geometric skeleton $\mathcal{S}_{geo}$ and text description $\mathcal{T}$:
\begin{equation}
    \mathcal{C}^{(0)} = \pi_{gen}(\mathcal{S}_{geo}, \mathcal{T})
\end{equation}

\paragraph{(2) Code Execution Agent ($\pi_{exec}$)}
This agent functions as a deterministic rendering engine, translating abstract code into visual pixel images without introducing stochastic noise:
\begin{equation}
    \mathcal{I}_{rec}^{(t)} = \pi_{exec}(\mathcal{C}^{(t)})
\end{equation}

\paragraph{(3) Hybrid Inspection Agent ($\pi_{insp}$)}
Acting as the Discriminator, this agent implements the \textbf{Visual Error Projection (VEP)} strategy. It compares $\mathcal{I}_{rec}^{(t)}$ with $\mathcal{I}_{obs}$ to bridge the modality gap.
\begin{itemize}
    \item \textbf{Metric Calculation:} It computes the Chamfer Distance (CD) and Hausdorff Distance (HD) to quantify edge alignment:
    \begin{equation}
        d_{CD}^{(t)}, d_{HD}^{(t)} = \text{Metric}(\text{Edge}(\mathcal{I}_{rec}^{(t)}), \text{Edge}(\mathcal{I}_{obs}))
    \end{equation}
    \item \textbf{Visual Error Projection:} Crucially, it projects numerical errors into a \textbf{Visual Difference Map} $\mathcal{M}_{err}^{(t)}$, enabling the model to see the mathematical divergence:
    \begin{equation}
        \mathcal{M}_{err}^{(t)} = \pi_{insp}(\mathcal{I}_{rec}^{(t)}, \mathcal{I}_{obs}, d_{CD}^{(t)}, d_{HD}^{(t)})
    \end{equation}
\end{itemize}

\paragraph{(4) Reflective Correction Agent ($\pi_{ref}$)}
This agent serves as the Optimizer. Observing the Visual Difference Map $\mathcal{M}_{err}^{(t)}$, it performs gradient-free optimization to generate an improved code version $\mathcal{C}^{(t+1)}$:
\begin{equation}
\begin{aligned}
    \mathcal{C}^{(t+1)} &= \pi_{ref}(\mathcal{C}^{(t)}, \mathcal{M}_{err}^{(t)}, \mathcal{S}_{geo}) \\
    &= \underset{\mathcal{C}}{\arg\max} \, P\left( \mathcal{C} \mid \mathcal{C}^{(t)}, \mathcal{M}_{err}^{(t)}, \mathcal{S}_{geo} \right)
\end{aligned}
\end{equation}

The loop terminates when error metrics fall below a convergence threshold $\epsilon$ or the iteration limit is reached.

\subsection{Optimization Objective}

Based on the problem definition, the global loss function $Q$ is defined as the weighted sum of three terms:
\begin{equation}
\begin{aligned}
    Q(\mathcal{C}, \mathcal{I}_{rec}) = \, &\alpha \cdot \mathcal{D}_{geo}(\mathcal{I}_{rec}, \mathcal{I}_{obs}) + \beta \cdot \mathcal{D}_{consist}(\mathcal{C}) \\
    &+ \gamma \cdot \mathcal{D}_{sem}(\mathcal{C}, \mathcal{T})
\end{aligned}
\end{equation}
where $\mathcal{D}_{geo}$ comprises CD and HD metrics for geometric fidelity, while $\mathcal{D}_{consist}$ and $\mathcal{D}_{sem}$ constrain grammatical correctness and semantic consistency, respectively.

\subsection{Illustrative Example}

To intuitively elucidate this multi-agent collaborative process, we present a solid geometry case study from Dataset A (refer to Figure~\ref{Main_img}). The task input comprises the observation $\mathcal{I}_{obs}$ and the descriptive text $\mathcal{T}$.

\textbf{Phase I: Geometric Skeleton Construction via Anchoring}

The process initiates in Phase I, where the \textbf{Geometric Extraction Agent ($\pi_{ext}$)} processes multimodal information in parallel (Figure~\ref{Main_img}, Part 1.2):
\begin{itemize}
    \item It parses high-level topological constraints $\mathcal{R}$ from text $\mathcal{T}$, identifying the structure as Cube $ABCD-A_1B_1C_1D_1$ and extracting relationships such as Points $E, F, G$ are midpoints.
    \item Simultaneously, it employs \textbf{Pixel-wise Anchoring Operators} to detect potential corners and intersections from $\mathcal{I}_{obs}$, generating the raw candidate set $\mathbf{P}_{raw}$. As shown in Figure~\ref{Main_img}, this initial set contains significant background noise.
\end{itemize}

Subsequently, the \textbf{Visual Verification Agent ($\pi_{ver}$)} intervenes. It performs semantic filtering on $\mathbf{P}_{raw}$ guided by $\mathcal{R}$, eliminating noise and explicitly grounding keypoints (e.g., vertex $A$, midpoint $E$) to precise pixel coordinates. This results in a clean geometric skeleton $\mathcal{S}_{geo}$ (Figure~\ref{Main_img}, Part 1.3).

\textbf{Phase II: Code Evolution via Visual Error Projection}

The grounded skeleton enters Phase II. The \textbf{Code Generation Agent ($\pi_{gen}$)} synthesizes the initial program, and the \textbf{Code Execution Agent ($\pi_{exec}$)} renders the first reconstruction $\mathcal{I}_{rec}^{(0)}$. At this stage, artifacts like missing segments or coordinate drift are common.

The system then enters the Inspection-Correction loop driven by the \textbf{Visual Error Projection (VEP)} mechanism:

The \textbf{Hybrid Inspection Agent ($\pi_{insp}$)} computes geometric metrics (CD/HD) and projects them into a \textbf{Visual Difference Map $\mathcal{M}_{err}^{(t)}$}. This map is submitted to the \textbf{Reflective Correction Agent ($\pi_{ref}$)} for diagnostic reasoning:
\begin{itemize}
    \item \textbf{Missing Structure:} If $\mathcal{I}_{obs}$ contains high-response regions absent in $\mathcal{I}_{rec}^{(t)}$ (e.g., a hot spot on segment $EF$), it triggers Geometric Completion.
    \item \textbf{Hallucination:} If the reconstruction shows high response where the original is blank, it triggers Redundancy Pruning.
    \item \textbf{Drift Calibration:} Partial overlap with center misalignment triggers Coordinate Fine-tuning.
    \item \textbf{Style Correction:} It validates attributes like line thickness and label placement.
\end{itemize}

As illustrated in Figure~\ref{Main_img} (Part 2.2), the VEP map reveals a positional offset of point $F$ (highlighted by the overlap discrepancy). Observing this visual cue, the Correction Agent fine-tunes point $F$'s parameters. This cycle iterates until the Hausdorff Distance $d_{HD}$ falls below $\epsilon$, yielding the final optimized code $\mathcal{C}^*$.
\section{Geo-Code Dataset and GeocodeLM}
\label{sec:dataset_and_model}

To bridge the critical gap in existing multimodal resources—specifically the lack of alignable code representations—we introduce \textbf{Geo-Code}, a high-fidelity benchmark explicitly designed for geometric program synthesis. Furthermore, to validate the efficacy of this dataset, we propose \textbf{Geo-CodeLM}, a specialized model fine-tuned on these high-quality annotations.

\subsection{The Geo-Code Dataset}
We leverage diverse open-source geometric sources—including \textit{AuxSolidMath}, \textit{GeoQA}, \textit{GeoSketch}, and \textit{MathVerse}—as the foundational data. Given the prohibitive cost of expert annotation and the stringent requirement for geometric precision, we prioritized data quality over quantity, constructing a compact yet dense instruction-tuning corpus.

\textbf{Construction Pipeline and Quality Control:}
The construction follows a rigorous Synthesis-Filtering-Verification pipeline:
\textbf{Automated Topological Filtering:} We recognize that traditional pixel-level metrics fail to capture structural integrity. Thus, we utilize \textbf{Chamfer Distance (CD)} and large-model visual verification to quantify the alignment between reconstructed primitives and the original input. By enforcing a stringent threshold (CD $< 10$ pixels), we filter out samples with topological deviations.
\textbf{Expert-in-the-loop Verification:} To ensure logical correctness, three mathematics professionals manually reviewed the filtered samples. This labor-intensive process ensures that every entry serves as a Golden Standard.

Each valid entry is structured as a \textbf{Quadruplet}: $\langle$Input Image, Geometric Attributes (JSON), Executable Code, Rendered Image$\rangle$. This meticulous process yielded a core set of \textbf{1,510 high-quality sample pairs}. While the scale is concise, each sample contains dense geometric reasoning paths, making it highly effective for instruction tuning.

\subsection{Geo-CodeLM}
Based on the constructed Geo-Code dataset, we developed \textbf{Geo-CodeLM} to establish a strong baseline for the Image-to-Geometry-Code task.We selected \textbf{Qwen3-4B-Instruction} as our basemodel.
We fine-tuned the model using the 1,510 high-quality pairs. Despite the dataset size, the high signal-to-noise ratio allowed for efficient convergence. The training was conducted on a single NVIDIA A800 (80GB) GPU. We employed a full-parameter fine-tuning strategy with a batch size of 8 and a learning rate of $1 \times 10^{-5}$. To ensure reproducibility, we fixed the random seed and utilized a greedy decoding strategy during the inference phase.

The performance of Geo-CodeLM, as detailed in the Experiment section, demonstrates that a small amount of high-quality data can significantly activate the model's domain-specific capabilities.

\section{Experiment}

\begin{table*}[t]
\centering
\caption{Average experimental results across four benchmark datasets. Methods are categorized into four logical groups. \color{red}Red backgrounds\color{black} indicate the best average performance. $\uparrow/\downarrow$ denotes the improvement relative to the corresponding baseline (Gemini for Geo-Code, Qwen for GeoCodeLM).}
\label{tab:average_results_advanced}
\begin{tabular}{lccccccc}
\toprule
 & \multicolumn{4}{c}{Visual Consistency (Average)} & \multicolumn{3}{c}{Pixel Consistency (Average)} \\
\cmidrule(lr){2-5} \cmidrule(lr){6-8}
Method & SC ($\uparrow$) & PP ($\uparrow$) & SCp ($\uparrow$) & LO ($\uparrow$) & SSIM ($\uparrow$) & HD ($\downarrow$) & CD ($\downarrow$) \\
\midrule
\multicolumn{8}{c}{--- \textbf{Reference} ---} \\
Original Image & 100 & 100 & 100 & 100 & 1 & 0 & 0 \\
\midrule
\multicolumn{8}{c}{--- \textbf{Specialized Methods} ---} \\
FigCodifier & 72.87 & 89.70 & 79.14 & 79.12 & 0.8325 & 225.59 & 66.13 \\
MatPlotCode & 82.81 & 93.75 & 83.52 & 85.55 & 0.8106 & 157.27 & 39.27 \\
V-Thinker & 88.79 & 96.25 & 93.17 & 90.79 & 0.8480 & 112.83 & 22.45 \\
\midrule
\multicolumn{8}{c}{--- \textbf{General MLLM Baselines} ---} \\
Qwen-4B-Instruction & 52.35 & 77.48 & 56.57 & 60.89 & 0.8163 & 277.44 & 63.24 \\
gemini-3-flash-preview & 85.38 & 98.77 & 92.42 & 91.31 & 0.8400 & 185.77 & 47.26 \\
\midrule
\multicolumn{8}{c}{--- \textbf{Ours} ---} \\
Geo-CodeLM & 81.95 \scalebox{0.7}{\textcolor{red}{$\uparrow$29.6}} & 92.13 \scalebox{0.7}{\textcolor{red}{$\uparrow$14.6}} & 82.93 \scalebox{0.7}{\textcolor{red}{$\uparrow$26.3}} & 85.23 \scalebox{0.7}{\textcolor{red}{$\uparrow$24.3}} & \cellcolor{red!20}0.8597 \scalebox{0.7}{\textcolor{red}{$\uparrow$.04}} & 116.13 \scalebox{0.7}{\textcolor{red}{$\downarrow$161}} & 19.97 \scalebox{0.7}{\textcolor{red}{$\downarrow$43.2}} \\
Geo-code & \cellcolor{red!20}97.97 \scalebox{0.7}{\textcolor{red}{$\uparrow$12.5}} & \cellcolor{red!20}98.90 \scalebox{0.7}{\textcolor{red}{$\uparrow$0.1}} & \cellcolor{red!20}98.00 \scalebox{0.7}{\textcolor{red}{$\uparrow$5.5}} & \cellcolor{red!20}98.04 \scalebox{0.7}{\textcolor{red}{$\uparrow$6.7}} & 0.8456 \scalebox{0.7}{\textcolor{red}{$\uparrow$.01}} & \cellcolor{red!20}95.90 \scalebox{0.7}{\textcolor{red}{$\downarrow$89.8}} & \cellcolor{red!20}13.06 \scalebox{0.7}{\textcolor{red}{$\downarrow$34.2}} \\
\bottomrule
\multicolumn{8}{l}{\footnotesize SC: Structural Consistency, PP: Points Presence, SCp: Segments Completeness, LO: LLM Overall Score (maximum score is 100).} \\
\multicolumn{8}{l}{\footnotesize SSIM: Structural Similarity Index Measure (maximum score is 1).} \\
\multicolumn{8}{l}{\footnotesize HD: Hausdorff Distance, CD: Chamfer Distance (lower values closer to 0 indicate better performance).} \\
\end{tabular}
\end{table*}

To comprehensively evaluate the effectiveness of the Geo-code framework in the task of geometric reconstruction, we design three core research studies:
\begin{enumerate*}[label=(\arabic*)]
    \item \textbf{Baseline Comparative Study}: Evaluating the performance of Geo-code against state-of-the-art image-to-code methods across four benchmark datasets: \textit{MathVerse}, \textit{AuxSolidMath}, \textit{GeoQA}, and \textit{GeoSketch}.
    \item \textbf{Downstream Reasoning Evaluation}: Comparing the performance of images reconstructed by different methods in multimodal geometric reasoning tasks (see Section \ref{6.2}).
    \item \textbf{Ablation Study}: Deconstructing the core components of the framework to investigate the contribution of each module (see Section \ref{6.3}).
\end{enumerate*}

\subsection{Comparative Study and Results Analysis}
\label{6.1}
\subsubsection{Experimental Setup}

\textbf{Datasets Selection:} The experiments cover a diverse range of geometric challenges. \textbf{AuxSolidMath} is a pure solid geometry dataset with complex topological relationships and no auxiliary annotations, serving to test 3D structural fidelity. \textbf{GeoSketch} and \textbf{GeoQA} are planar geometry datasets featuring standard annotations like coordinate systems and perpendicular markers, used to evaluate the reconstruction of common geometric primitives. \textbf{MathVerse} contains function plots and non-geometric structures to verify the framework's generalization capability.

\textbf{Evaluation Metrics:} We assess the quality of reconstructed images from two dimensions:
\begin{itemize}
    \item \textbf{Visual Semantic Fidelity}: We utilize \textit{Gemini 3-pro} as an evaluator to score images based on \textit{Structural Consistency} (SC), \textit{Point Positioning} (PP) accuracy, and \textit{Segment/Arc Precision} (SCp). These are aggregated into a comprehensive \textit{Layout Score} (LO).
    \item \textbf{Pixel-level Precision}: On a standardized 1000px canvas, we calculate the \textit{Structural Similarity Index} (SSIM), \textit{Chamfer Distance} (CD), and \textit{Hausdorff Distance} (HD) between the original and reconstructed images.
\end{itemize}

\textbf{Baseline Methods:} We compare our framework against: (1) specialized fine-tuned models for code generation, \textit{MatPlotCode} (30B) and \textit{FigCodifier} (30B); and (2) \textit{Gemini 3-flash-preview} as a Zero-shot baseline for direct image-to-code generation.

\subsubsection{Results and Discussion}

The experimental results are illustrated in Table~\ref{tab:average_results_advanced}. Table~\ref{tab:average_results_advanced} presents the average scores across datasets (refer to the Appendix for detailed breakdowns).

\textbf{Analysis of the Geo-code Framework:}
Geo-code significantly outperforms all baseline methods across all metrics. In the \textbf{Visual Semantic} dimension, it achieves an average LO of 98.04, an improvement of +6.73 over the base model, demonstrating high fidelity in restoring geometric topology. In the \textbf{Pixel-level} dimension, the SSIM reaches 84.56, indicating high stylistic consistency with the ground truth.


Regarding distance metrics, Geo-code achieves a mean CD of 13.06 pixels, suggesting minimal average deviation between the reconstructed and original point sets. Notably, regarding the \textbf{Hausdorff Distance} (HD), which captures the maximum local error (worst-case scenario), Geo-code achieves 95.9 pixels. \textbf{While this absolute value reflects the inherent challenge of pixel-perfect alignment, it represents a substantial improvement compared to baselines (e.g., FigCodifier's 225.59 and Qwen's 277.44). This demonstrates that Geo-code effectively suppresses catastrophic structural collapse, keeping even the most deviant outliers under significantly better control than end-to-end models.}

\textbf{Performance Gains of Geo-codeLM:}
The fine-tuned \textit{Geo-codeLM} exhibits enhanced geometric perception. The visual metrics SC and LO improve by +29.6 and +24.3, respectively, reflecting the model's mastered logic of structural composition. In terms of pixel precision, the CD drops from 66.24 to 19.97. This leap in accuracy demonstrates \textit{Geo-codeLM}'s ability to deeply recognize geometric skeletons and image metadata.
A critical finding is that \textit{Geo-codeLM} outperforms \textit{FigCodifier} and \textit{MatPlotCode} in pixel-level metrics, despite the latter being fine-tuned on over 1 million samples. This strongly supports our hypothesis: \textbf{In geometric inverse tasks, the quality of geometric logic in training data is far more critical than sheer volume.} Our high-quality synthesized dataset provides precise geometric anchors that effectively guide the model in establishing a mapping from pixels to geometric programs.

\begin{figure*}[!t]
    \centering
    \begin{subfigure}{0.24\linewidth}
        \centering
        \includegraphics[width=\linewidth]{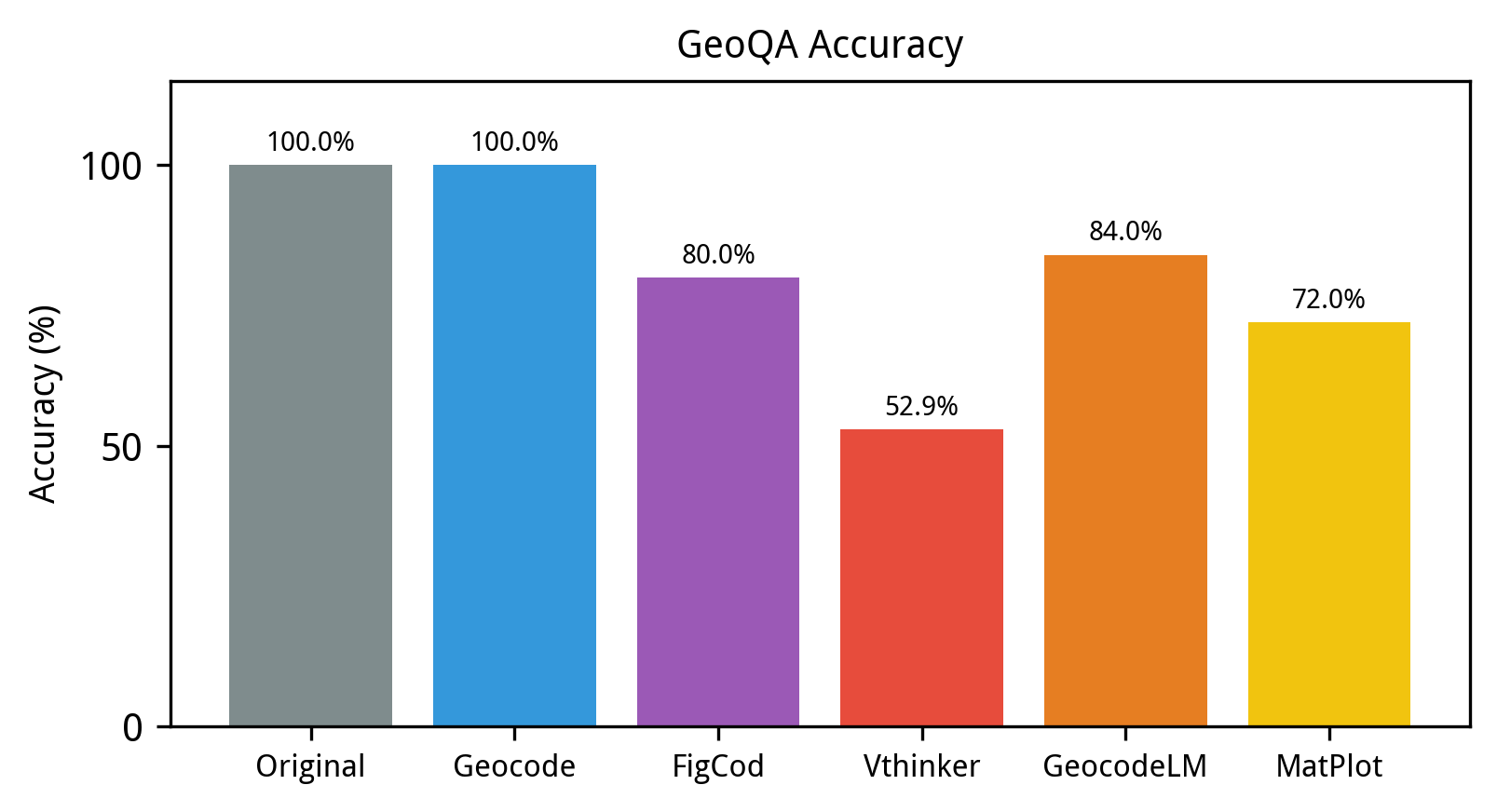}
        \caption{GeoQA}
        \label{fig:geoqa}
    \end{subfigure}
    \hfill
    \begin{subfigure}{0.24\linewidth}
        \centering
        \includegraphics[width=\linewidth]{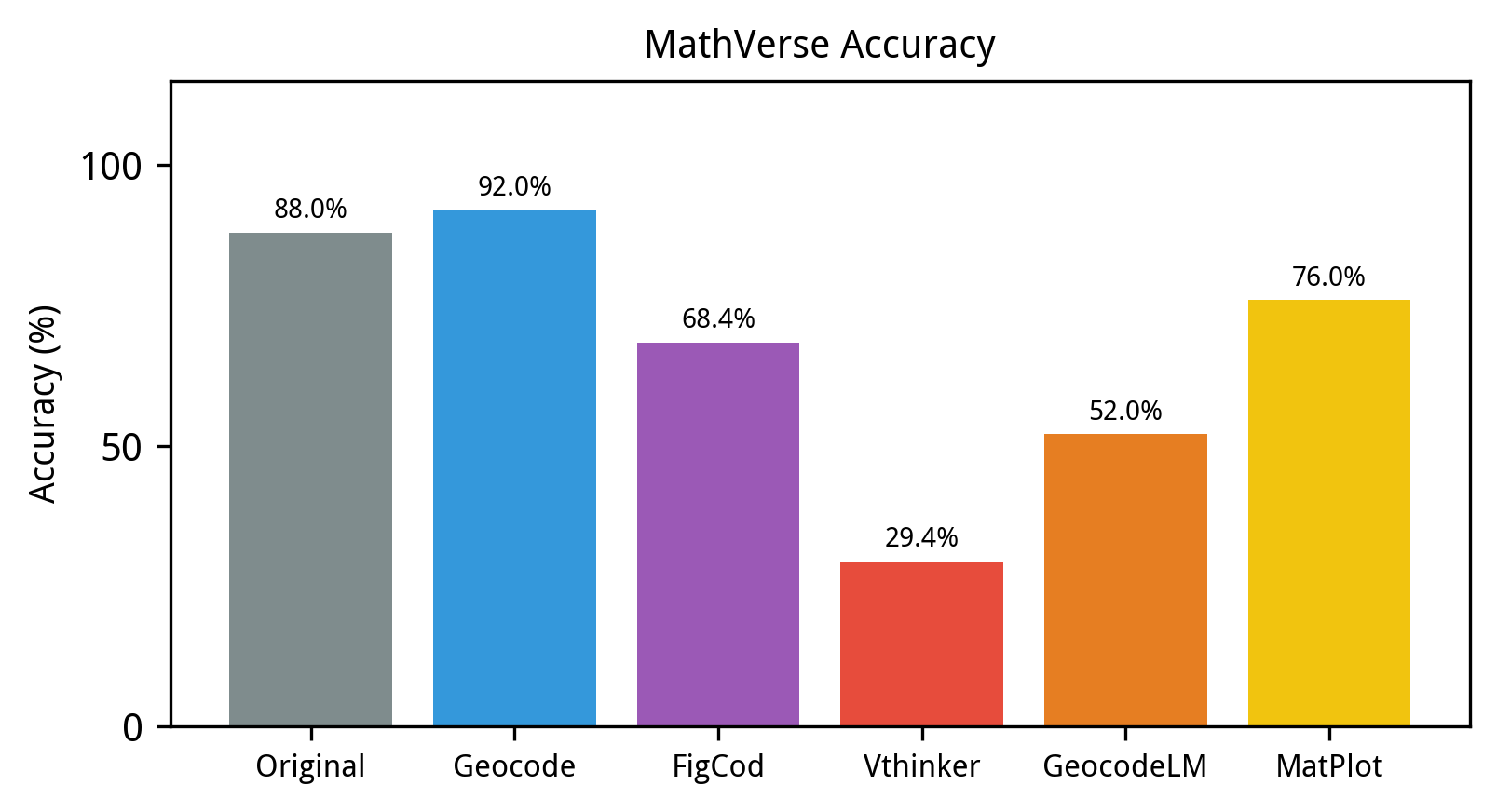}
        \caption{MathVerse}
        \label{fig:mathverse}
    \end{subfigure}
    \hfill
    \begin{subfigure}{0.24\linewidth}
        \centering
        \includegraphics[width=\linewidth]{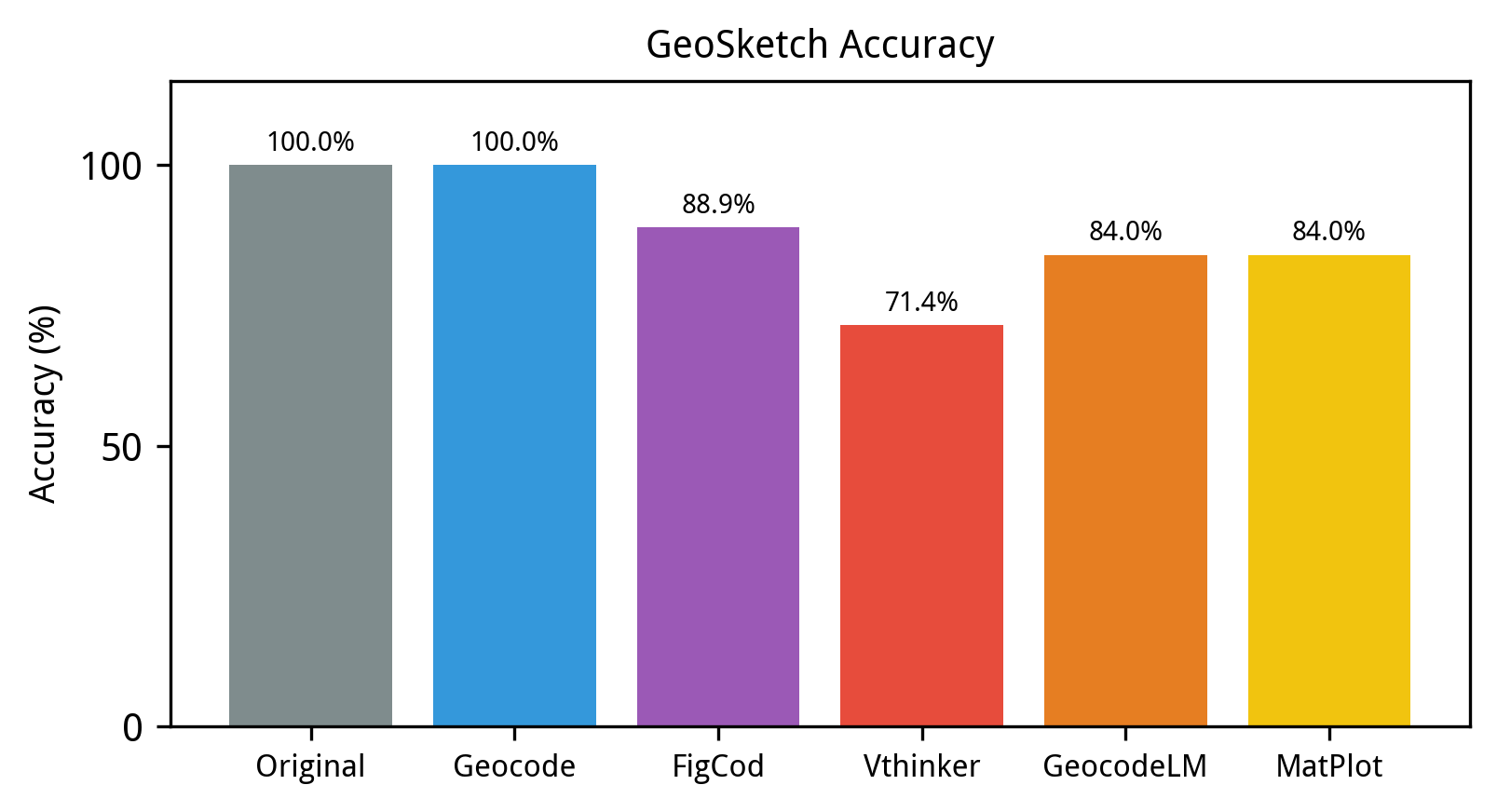}
        \caption{GeoSketch}
        \label{fig:geosketch}
    \end{subfigure}
    \hfill
    \begin{subfigure}{0.24\linewidth}
        \centering
        \includegraphics[width=\linewidth]{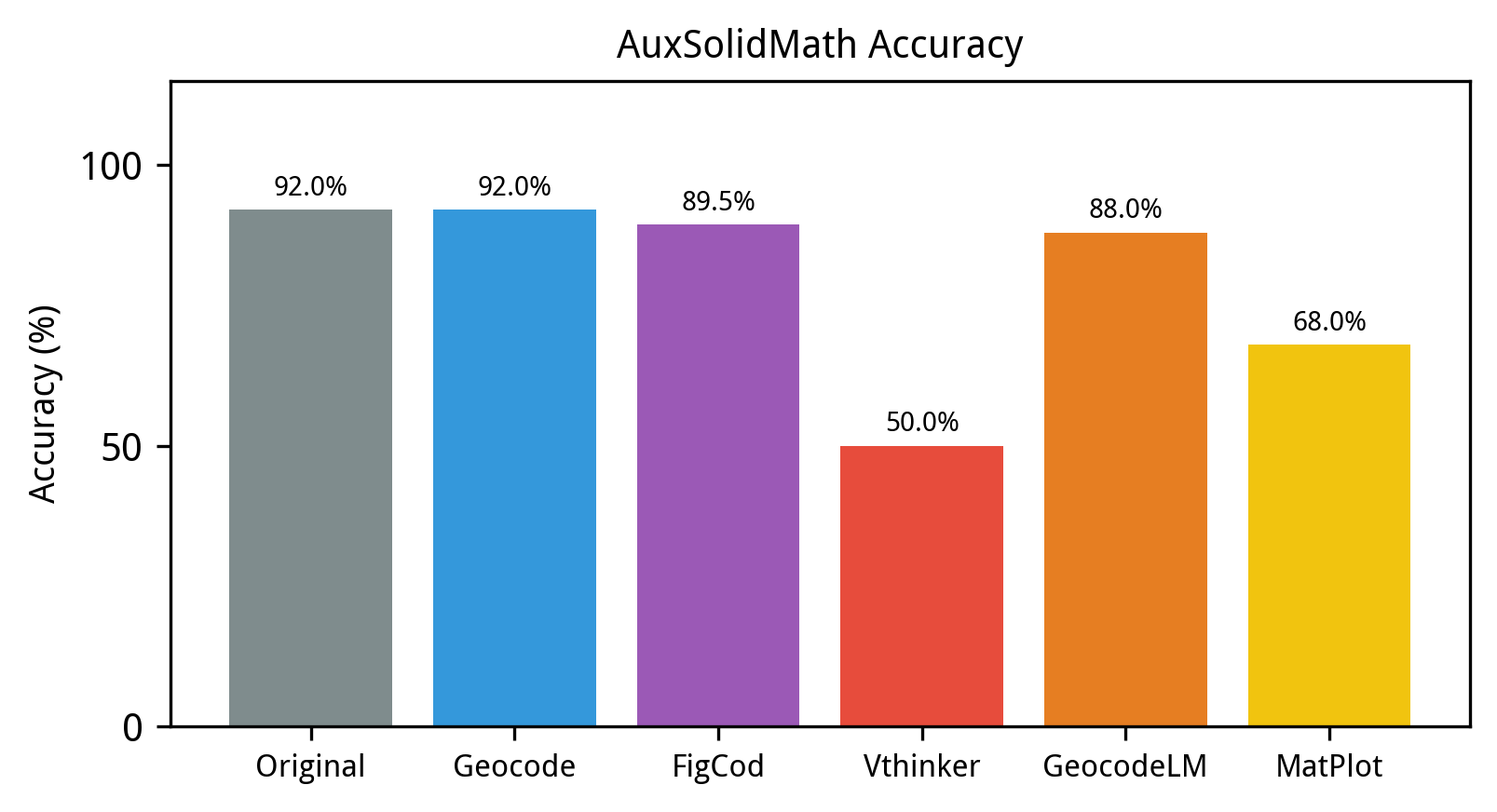}
        \caption{AuxSolidMath}
        \label{fig:auxsolidmath}
    \end{subfigure}
    
    \caption{Detailed inference accuracy for each individual dataset across four benchmarks.}
    \label{fig:detailed_accuracy_single_row}
\end{figure*}

\begin{table*}[h]
\centering
\caption{Ablation study on AuxSolidMath dataset. Improvements/regressions are calculated relative to the Gemini baseline. \color{red}Red arrows\color{black} indicate performance gains, while \color{blue}blue arrows\color{black} indicate regressions.}
\label{tab:ablation_study_advanced}
\begin{tabular}{lccccccc}
\toprule
 & \multicolumn{4}{c}{Visual Consistency} & \multicolumn{3}{c}{Pixel Consistency} \\
\cmidrule(lr){2-5} \cmidrule(lr){6-8}
Method & SC ($\uparrow$) & PP ($\uparrow$) & SCp ($\uparrow$) & LO ($\uparrow$) & SSIM ($\uparrow$) & HD ($\downarrow$) & CD ($\downarrow$) \\
\midrule
gemini-3-flash-preview & 85.38 & 98.77 & 92.42 & 91.31 & 0.8400 & 185.77 & 47.26 \\
\midrule
+ Phase 1 & \cellcolor{red!20}98.74 \scalebox{0.7}{\textcolor{red}{$\uparrow$13.3}} & 99.19 \scalebox{0.7}{\textcolor{red}{$\uparrow$0.4}} & \cellcolor{red!20}99.21 \scalebox{0.7}{\textcolor{red}{$\uparrow$6.8}} & \cellcolor{red!20}99.07 \scalebox{0.7}{\textcolor{red}{$\uparrow$7.8}} & \cellcolor{red!20}0.8667 \scalebox{0.7}{\textcolor{red}{$\uparrow$.03}} & 61.78 \scalebox{0.7}{\textcolor{red}{$\downarrow$124}} & 8.00 \scalebox{0.7}{\textcolor{red}{$\downarrow$39.3}} \\
+ Phase2(Geo-code) & 96.68 \scalebox{0.7}{\textcolor{red}{$\uparrow$11.3}} & \cellcolor{red!20}99.74 \scalebox{0.7}{\textcolor{red}{$\uparrow$1.0}} & 97.68 \scalebox{0.7}{\textcolor{red}{$\uparrow$5.3}} & 97.84 \scalebox{0.7}{\textcolor{red}{$\uparrow$6.5}} & 0.8331 \scalebox{0.7}{\textcolor{blue}{$\downarrow$.01}} & \cellcolor{red!20}60.56 \scalebox{0.7}{\textcolor{red}{$\downarrow$125}} & \cellcolor{red!20}6.41 \scalebox{0.7}{\textcolor{red}{$\downarrow$40.9}} \\
\bottomrule
\end{tabular}
\end{table*}

\subsection{Downstream Multimodal Reasoning Evaluation}
\label{6.2}

To further validate the utility of reconstructed images in practical applications, we conducted downstream multimodal reasoning experiments. Specifically, we employed \textit{Gemini-3-flash} as the core reasoning engine to solve geometry problems based on images generated by six methods: \textbf{Original Images}, \textbf{Geo-code}, \textbf{Geo-codeLM}, \textbf{Vthink}, \textbf{MatPlotCode}, and \textbf{FigCodifier}. The results are illustrated in Figure~\ref{fig:detailed_accuracy_single_row}.

\subsubsection{Results and Semantic Fidelity Analysis}

\textbf{Fidelity and Enhancement of Geo-code:} 
The original images serve as the performance upper bound, achieving 100\% accuracy on \textit{GeoSketch} and \textit{GeoQA}, and 92\% on \textit{MathVerse} and \textit{AuxSolidMath}. Remarkably, \textit{Geo-code} not only maintains this high fidelity but also \textbf{outperforms the original images by +4\% on MathVerse}. This indicates that our framework effectively performs \textit{geometric denoising and logical clarification}. By converting noisy raster pixels into precise geometric primitives, \textit{Geo-code} enhances the reasoning engine's ability to capture critical features.

\textbf{Performance Degradation in Baselines:} 
Standard methods like \textit{FigCodifier} and \textit{MatPlotCode} exhibit significant accuracy drops. While \textit{FigCodifier} shows relative stability on \textit{AuxSolidMath} (2.5\% loss), its accuracy plummets by nearly 20\% on \textit{MathVerse}, indicating that these models struggle to preserve semantic consistency in complex multimodal scenarios.
A key finding lies in the performance of the \textbf{Vthink} method. Although Vthink exhibits superior pixel-level metrics in Figure~\ref{fig:detailed_accuracy_single_row} (e.g., leading \textit{FigCodifier} and \textit{MatPlotCode} in CD by 43.68 and 16.82, respectively), its reasoning accuracy is substantially lower. On \textit{GeoQA}, its accuracy is 31.1\% lower than \textit{FigCodifier} and 19.1\% lower than \textit{MatPlotCode}. 

\textbf{This paradox underscores a fundamental insight: pixel-level similarity does not equate to geometric correctness.} While \textit{Vthink} produces images that are visually similar to the ground truth, they likely contain latent topological errors or incorrect mathematical expressions. This contrast highlights the necessity of the \textit{Geo-code} approach, which leverages code-based representations to ensure both visual fidelity and rigorous geometric logic.

\subsection{Ablation Study}
\label{6.3}

To quantitatively evaluate the contribution of each core component within the \textit{Geo-code} framework, we conducted an ablation study. The results, summarized in Table~\ref{tab:ablation_study_advanced}, demonstrate that every Phase of the framework contributes positively to the final reconstruction quality, exhibiting significant synergy.

\subsubsection{Impact of Phase 1}
Experimental results indicate that \textbf{Phase 1} (Qualitative Constraint Discovery and Hybrid Keypoint Extraction) provides the most substantial performance gain. By establishing a global geometric skeleton and identifying key topological constraints, this Phase defines the logical boundaries of the solution space. Without the structural guidance provided in Phase 1, the model is prone to topological failures in complex scenarios. Thus, Phase 1 is the primary contributor to ensuring the structural correctness of the reconstruction.


\subsubsection{Impact of Phase 2}
\textbf{Phase 2} (Reconstruction-based Iterative Evolution) focuses on rectifying coordinate drifts to achieve \textit{pixel-level precision}. As shown in Table~\ref{tab:ablation_study_advanced}, while the introduction of Phase 2 causes slight fluctuations in high-level semantic scores (e.g., SC and LO) due to the sensitivity of MLLM evaluators, it significantly improves the physical alignment metrics. 

Specifically, \textbf{Phase 2 reduces the Chamfer Distance (CD) from 8.00 to 6.41 and the Hausdorff Distance (HD) from 61.78 to 60.56.} This confirms that the feedback loop effectively fine-tunes the geometric parameters, pulling the reconstructed primitives closer to the ground truth coordinates. While Phase 1 determines the correct topology, Phase 2 is indispensable for minimizing quantization errors and achieving high-precision alignment.


\section{Conclusion}
In this paper, we presented \textbf{Geo-code}, a novel multi-agent-based inverse programming framework that bridges the gap between geometric vision and logic.  By decoupling anchoring-based geometric modeling from VEP-guided code evolution, our framework effectively addresses the long-standing challenges of structural distortion and detail loss in geometric reconstruction. The introduction of a synthesis-rendering-validation feedback loop ensures high visual consistency and mathematical rigor. Experimental results demonstrate that images reconstructed via Geo-code achieve performance parity with original images in complex multimodal reasoning tasks, confirming the framework's robustness. Furthermore, by open-sourcing the Geo-code dataset and the fine-tuned \textit{Geo-codeLM} model, we provide a solid foundation and valuable resources to accelerate future research in the intersection of inverse graphics and geometric reasoning.

\section{Impact Statement}
This paper presents work whose goal is to advance the field of Machine Learning. There are many potential societal consequences of our work, none which we feel must be specifically highlighted here.

\clearpage

\bibliographystyle{icml2026}
\bibliography{ref}

\clearpage

\appendix

\section{Reconstructed Image Comparison}

\begin{figure*}[t]
    \centering
    \begin{subfigure}{0.23\textwidth}
        \centering
        \includegraphics[width=\linewidth]{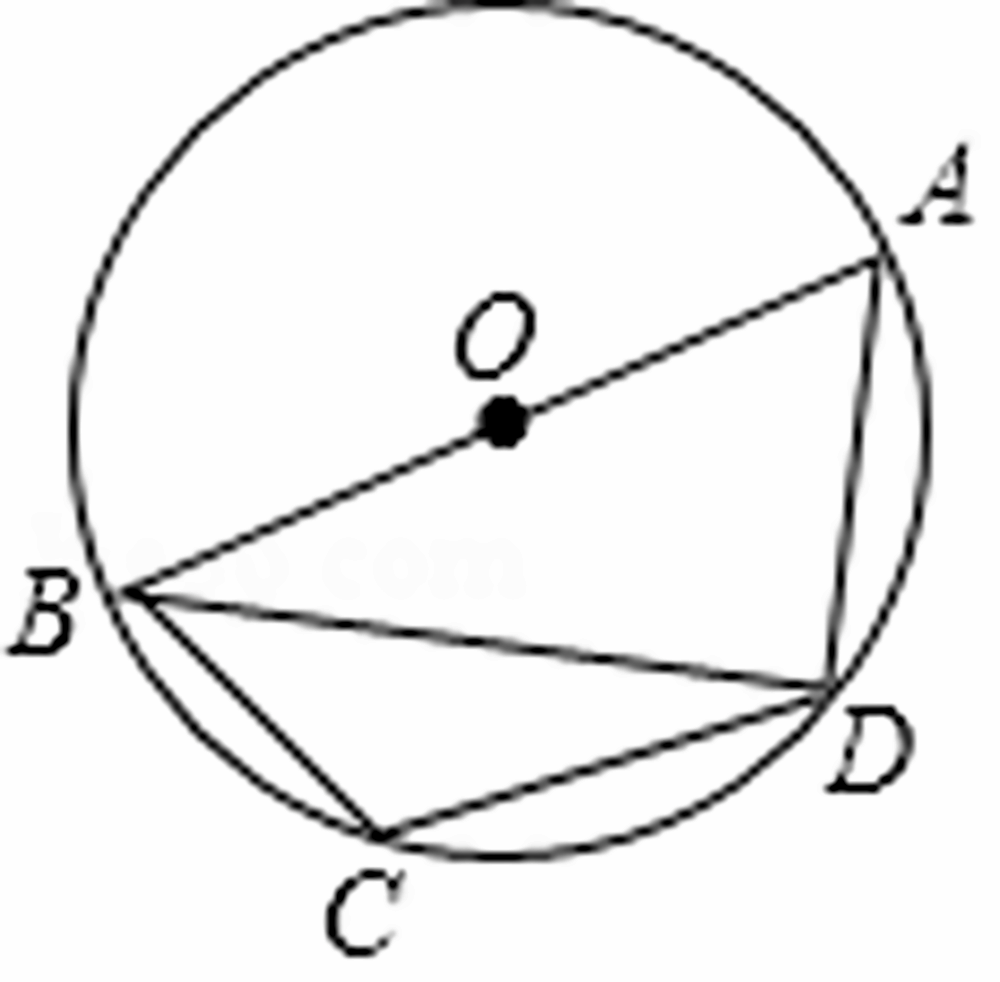}
        \subcaption{origin}
        \label{subfig:origin_157}
    \end{subfigure}
    \hfill
    \begin{subfigure}{0.23\textwidth}
        \centering
        \includegraphics[width=\linewidth]{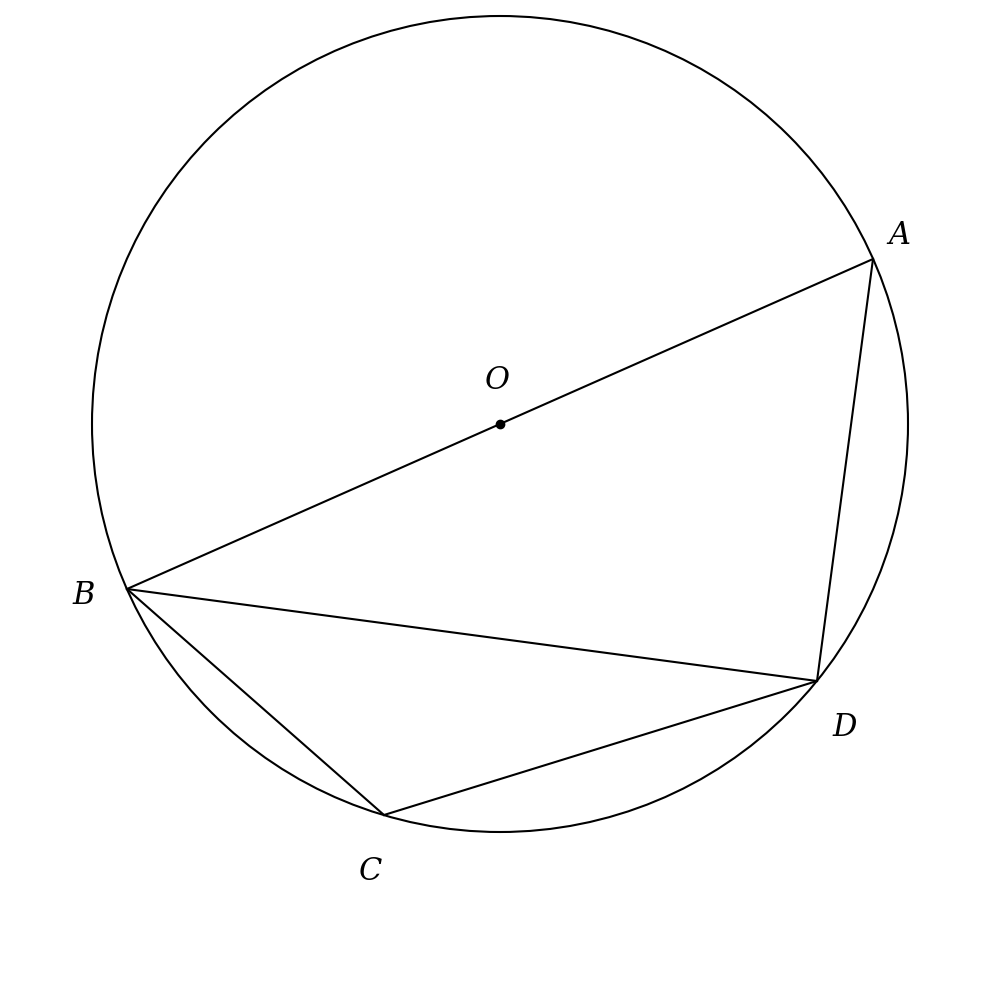}
        \subcaption{Geocode}
        \label{subfig:Geocode_157}
    \end{subfigure}
    \hfill
    \begin{subfigure}{0.23\textwidth}
        \centering
        \includegraphics[width=\linewidth]{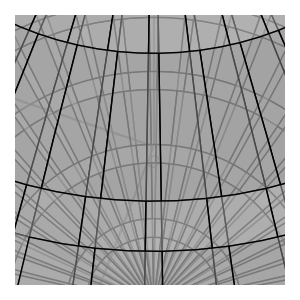}
        \subcaption{MatPlotCode}
        \label{subfig:math_157}
    \end{subfigure}
    \hfill
    \begin{subfigure}{0.23\textwidth}
        \centering
        \includegraphics[width=\linewidth]{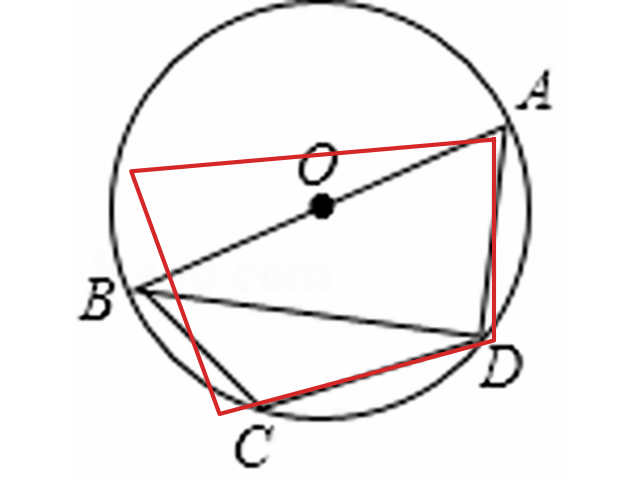}
        \subcaption{V-Thinker}
        \label{subfig:vthink_157}
    \end{subfigure}
    \caption{Comparison of Reconstructed Images (GeoQA No. 157)}
    \label{157}
\end{figure*}

\begin{figure*}[t]
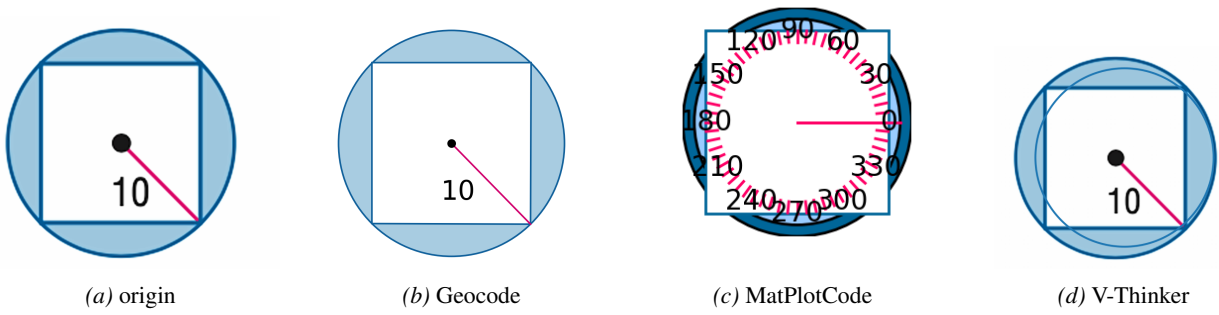
  
    \centering  
    
    \begin{subfigure}{0.23\textwidth}
        \centering
        \includegraphics[width=\linewidth]{Img/origin.png} 
        \subcaption{origin}
        \label{subfig:origin}
    \end{subfigure}
    \hfill 
    \begin{subfigure}{0.23\textwidth}
        \centering
        \includegraphics[width=\linewidth]{Img/Geocode.png}
        \subcaption{Geocode}
        \label{subfig:Geocode}
    \end{subfigure}
    \hfill
    \begin{subfigure}{0.23\textwidth}
        \centering
        \includegraphics[width=\linewidth]{Img/Math.png}
        \subcaption{MatPlotCode}
        \label{subfig:math}
    \end{subfigure}
    \hfill
    \begin{subfigure}{0.23\textwidth}
        \centering
        \includegraphics[width=\linewidth]{Img/Vthink.png}
        \subcaption{V-Thinker}
        \label{subfig:vthink}
    \end{subfigure}

    \caption{Reconstructed Images Comparison (GeoQA No.266)}
    \label{266}
\end{figure*}

\begin{figure*}[t]
    \centering
    \begin{subfigure}{0.23\textwidth}
        \centering
        \includegraphics[width=\linewidth]{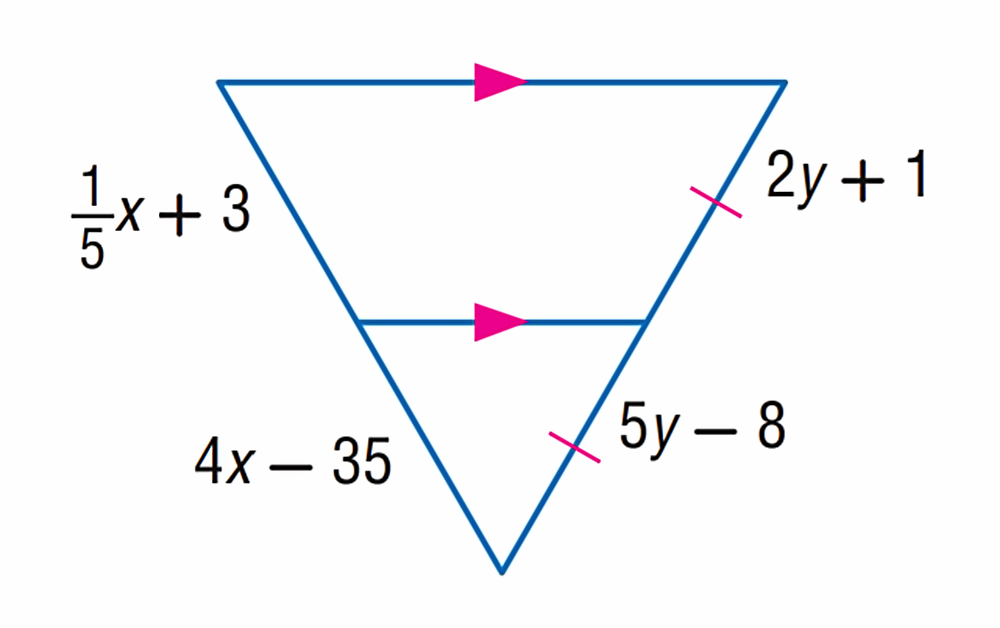}
        \subcaption{origin}
        \label{subfig:origin_300}
    \end{subfigure}
    \hfill
    \begin{subfigure}{0.23\textwidth}
        \centering
        \includegraphics[width=\linewidth]{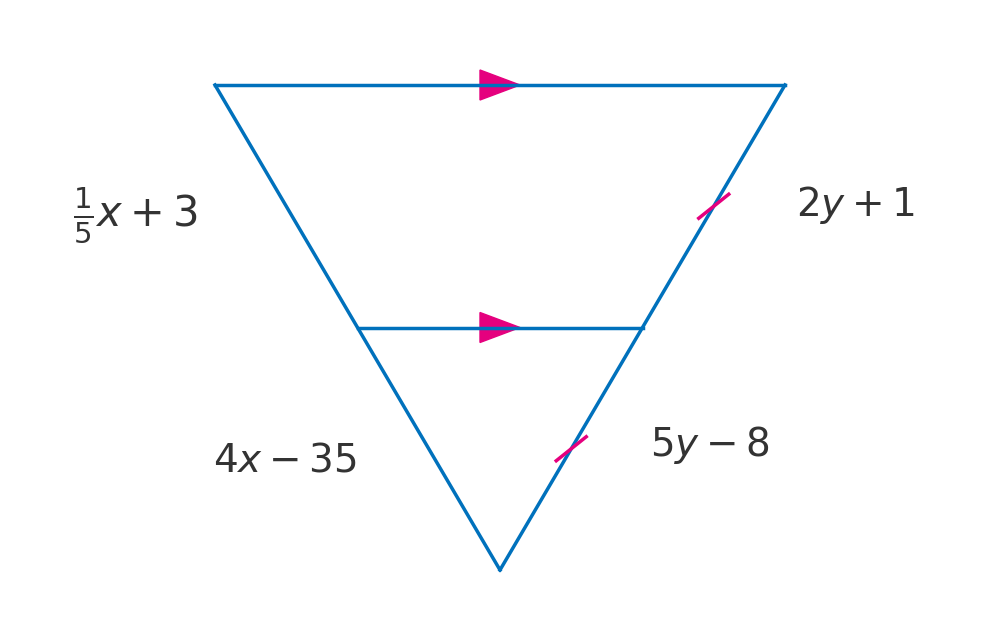}
        \subcaption{Geocode}
        \label{subfig:Geocode_300}
    \end{subfigure}
    \hfill
    \begin{subfigure}{0.23\textwidth}
        \centering
        \includegraphics[width=\linewidth]{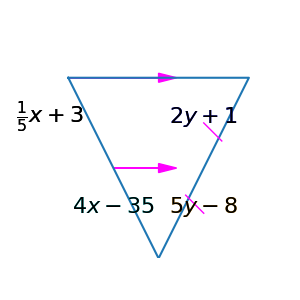}
        \subcaption{MatPlotCode}
        \label{subfig:math_300}
    \end{subfigure}
    \hfill
    \begin{subfigure}{0.23\textwidth}
        \centering
        \includegraphics[width=\linewidth]{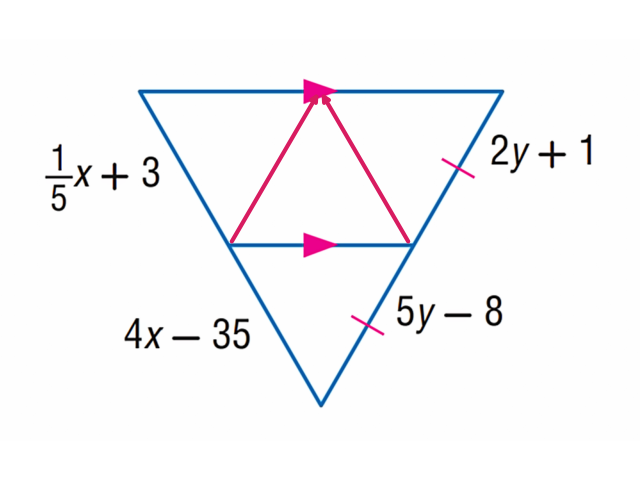}
        \subcaption{V-Thinker}
        \label{subfig:vthink_300}
    \end{subfigure}
    \caption{Comparison of Reconstructed Images (GeoQA No. 300)}
    \label{300}
\end{figure*}

\begin{figure*}[t]
    \centering
    \begin{subfigure}{0.23\textwidth}
        \centering
        \includegraphics[width=\linewidth]{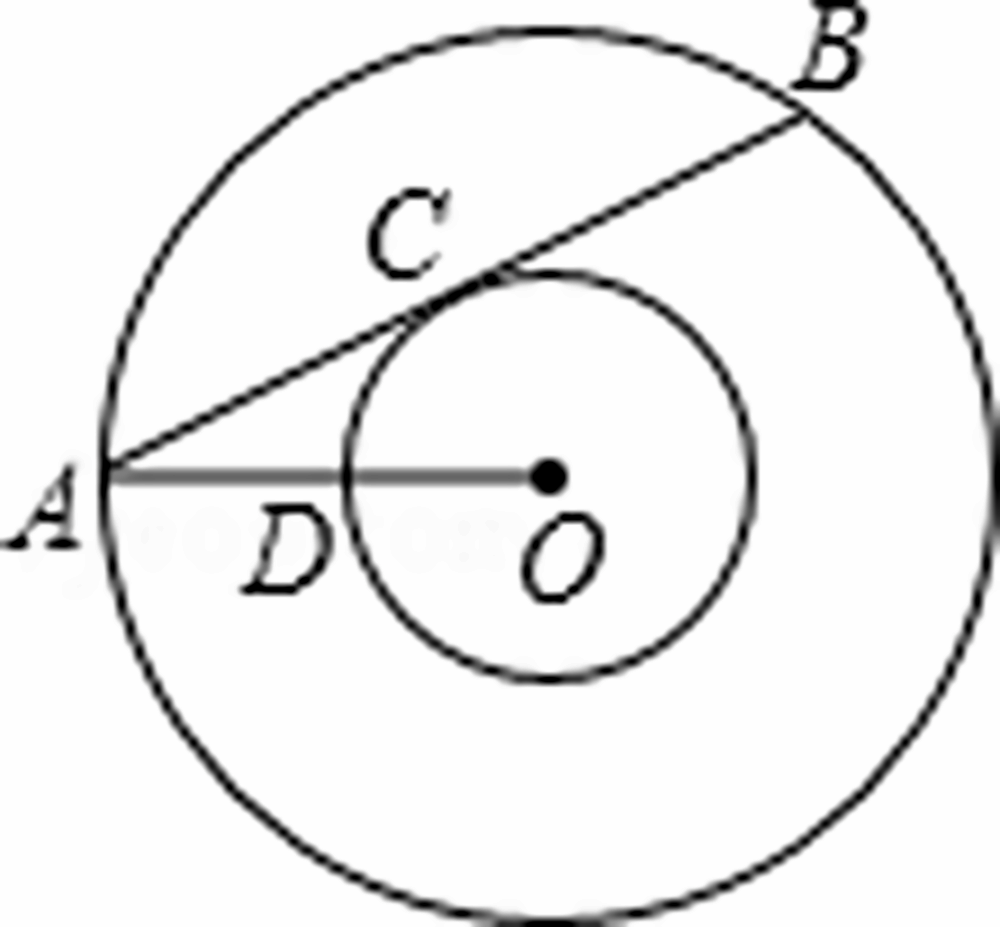}
        \subcaption{origin}
        \label{subfig:origin_498}
    \end{subfigure}
    \hfill
    \begin{subfigure}{0.23\textwidth}
        \centering
        \includegraphics[width=\linewidth]{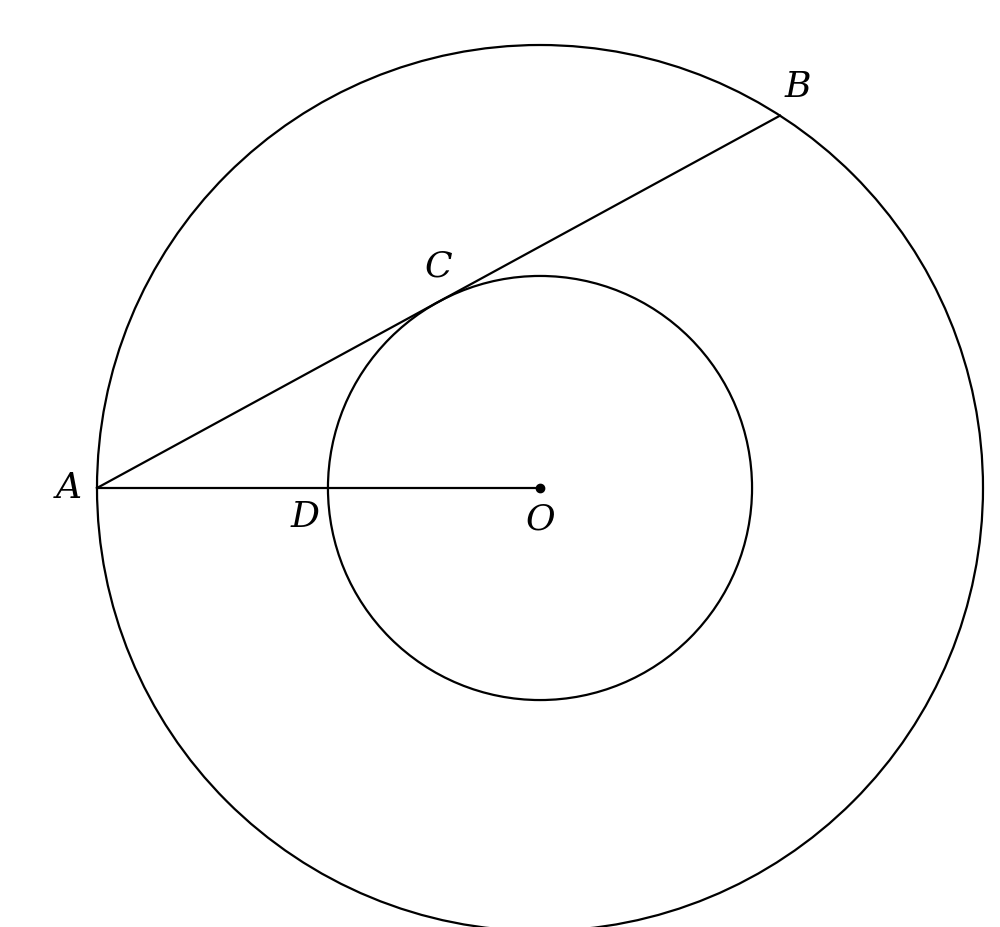}
        \subcaption{Geocode}
        \label{subfig:Geocode_498}
    \end{subfigure}
    \hfill
    \begin{subfigure}{0.23\textwidth}
        \centering
        \includegraphics[width=\linewidth]{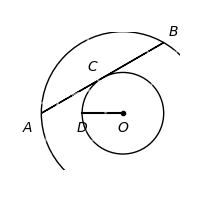}
        \subcaption{MatPlotCode}
        \label{subfig:math_498}
    \end{subfigure}
    \hfill
    \begin{subfigure}{0.23\textwidth}
        \centering
        \includegraphics[width=\linewidth]{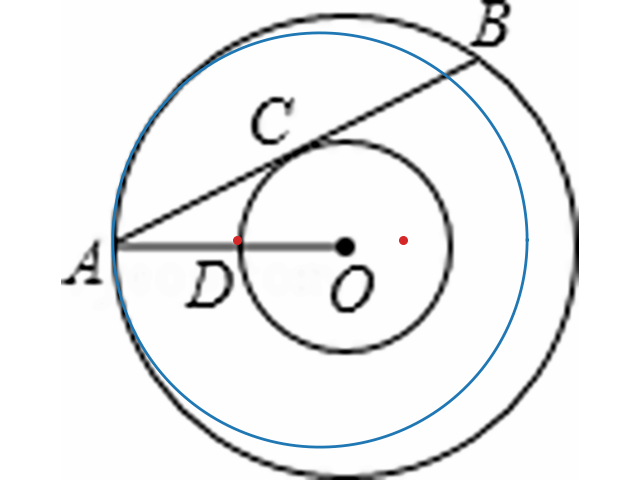}
        \subcaption{V-Thinker}
        \label{subfig:vthink_498}
    \end{subfigure}
    \caption{Comparison of Reconstructed Images (GeoQA No. 498)}
    \label{489}
\end{figure*}

As shown in Fig. \ref{157}, this figure presents a real-world example from the GeoQA dataset. Visual comparison of the images reveals that the Geocode method reconstructs the original image with the highest fidelity. In contrast, the image reconstructed by the MatPlotCode method is largely inconsistent with the original one: the original image consists of a circle with an embedded polygon, while MatPlotCode completely transforms it into a highly complex nesting of multiple shapes. This phenomenon indicates the occurrence of geometric hallucinations in the method. Although the result generated by Vthink bears a strong visual resemblance to the original image, an extraneous red shape is present in its output, which is inconsistent with the input.
As shown in Fig. \ref{266}, this figure presents a real-world example from the GeoQA dataset. Visual comparison of the images reveals that the Geocode method reconstructs the original image with the highest fidelity. In contrast, the MatPlotCode method suffers from over-reconstruction of the image structure and generates redundant annotations, a phenomenon that indicates the emergence of geometric hallucinations. Although the result generated by Vthink bears a strong visual resemblance to the original image, the associative relationship between the circle and the square is disrupted in its reconstructed output.
As shown in Fig. \ref{300}, this figure presents a real-world example from the GeoQA dataset. Visual comparison of the images reveals that the Geocode method reconstructs the original image with the highest fidelity. In contrast, a fracture appears in the middle of the image reconstructed by the MatPlotCode method: a blue line segment exists in the original image but disappears in the MatPlotCode’s output. The result generated by Vthink exhibits hallucinations, with multiple extraneous red line segments appearing in its reconstruction.
As shown in Fig. \ref{300}, this figure presents a real-world example from the GeoQA dataset. Visual comparison of the images reveals that the Geocode method reconstructs the original image with the highest fidelity. In contrast, the bottom-right portion of the image is missing in the reconstruction by the MatPlotCode method, which may be attributed to an error in canvas rendering. The result generated by Vthink exhibits hallucinations, with an extraneous red line segment in blue appearing in its reconstruction.

\section{complete comparative experiments}

\begin{table*}[t]
\centering
\caption{Comprehensive experimental results on visual and pixel consistency across all datasets. Red backgrounds indicate the best performance in each metric per dataset.}
\label{tab:consistency_results_highlighted}
\begin{tabular}{llccccccc}
\toprule
 & & \multicolumn{4}{c}{Visual Consistency} & \multicolumn{3}{c}{Pixel Consistency} \\
\cmidrule(lr){3-6} \cmidrule(lr){7-9}
Dataset & Method & SC ($\uparrow$) & PP ($\uparrow$) & SCp ($\uparrow$) & LO ($\uparrow$) & SSIM ($\uparrow$) & HD ($\downarrow$) & CD ($\downarrow$) \\
\midrule
\multirow{7}{*}{AuxSolidMath} & Original Image & - & - & - & - & - & - & - \\
\cmidrule(lr){2-9}
 & FigCodifier & 86.58 & 98.95 & 92.58 & 91.42 & 0.8295 & 198.31 & 54.06 \\
 & MatPlotCode & \cellcolor{red!20}98.17 & 99.17 & \cellcolor{red!20}98.67 & \cellcolor{red!20}98.39 & 0.8111 & 112.17 & 27.63 \\
\cmidrule(lr){2-9}
 & Qwen-4B-Instruction & 56.54 & 68.69 & 49.62 & 58.23 & 0.7815 & 285.90 & 66.89 \\
 & gemini-3-flash-preview & 85.38 & 98.77 & 92.42 & 91.31 & \cellcolor{red!20}0.8400 & 185.77 & 47.26 \\
\cmidrule(lr){2-9}
 & Geocode (v3.3\_pro) & 96.68 & \cellcolor{red!20}99.74 & 97.68 & 97.84 & 0.8331 & \cellcolor{red!20}60.56 & \cellcolor{red!20}6.41 \\
 & GeocodeLLM & 90.12 & 98.00 & 89.28 & 92.36 & 0.8314 & 92.11 & 15.13 \\
\midrule
\midrule
\multirow{7}{*}{GeoQA} & Original Image & - & - & - & - & - & - & - \\
\cmidrule(lr){2-9}
 & FigCodifier & 64.85 & 85.50 & 70.40 & 71.50 & 0.8404 & 220.86 & 78.78 \\
 & MatPlotCode & 74.42 & 94.47 & 74.58 & 79.53 & 0.8205 & 167.38 & 48.61 \\
\cmidrule(lr){2-9}
 & Qwen-4B-Instruction & 41.87 & 84.13 & 53.26 & 57.35 & 0.8329 & 271.87 & 84.29 \\
 & gemini-3-flash-preview & 85.38 & 98.77 & 92.42 & 91.31 & 0.8400 & 185.77 & 47.26 \\
\cmidrule(lr){2-9}
 & Geocode (v3.3\_pro) & \cellcolor{red!20}97.85 & \cellcolor{red!20}99.00 & \cellcolor{red!20}98.00 & \cellcolor{red!20}97.85 & \cellcolor{red!20}0.8562 & \cellcolor{red!20}73.52 & \cellcolor{red!20}14.41 \\
 & GeocodeLLM & 74.39 & 91.96 & 83.70 & 81.57 & 0.8531 & 116.01 & 23.67 \\
\midrule
\midrule
\multirow{7}{*}{GeoSketch} & Original Image & - & - & - & - & - & - & - \\
\cmidrule(lr){2-9}
 & FigCodifier & 80.56 & 97.78 & 89.61 & 88.72 & 0.8436 & 237.57 & 69.90 \\
 & MatPlotCode & 88.89 & \cellcolor{red!20}98.95 & 88.32 & 90.58 & 0.8118 & 161.51 & 38.85 \\
\cmidrule(lr){2-9}
 & Qwen-4B-Instruction & 53.83 & 87.83 & 65.65 & 67.09 & 0.8412 & 333.08 & 64.54 \\
 & gemini-3-flash-preview & 85.38 & 98.77 & 92.42 & 91.31 & 0.8400 & 185.77 & 47.26 \\
\cmidrule(lr){2-9}
 & Geocode (v3.3\_pro) & \cellcolor{red!20}99.35 & 97.60 & \cellcolor{red!20}97.95 & \cellcolor{red!20}98.20 & 0.8502 & 122.38 & \cellcolor{red!20}12.04 \\
 & GeocodeLLM & 92.10 & 98.57 & 90.57 & 93.62 & \cellcolor{red!20}0.8784 & \cellcolor{red!20}112.31 & 18.72 \\
\midrule
\midrule
\multirow{7}{*}{MathVerse} & Original Image & - & - & - & - & - & - & - \\
\cmidrule(lr){2-9}
 & FigCodifier & 59.47 & 76.58 & 63.95 & 64.84 & 0.8166 & 245.61 & 61.79 \\
 & MatPlotCode & 69.75 & 82.40 & 72.50 & 73.70 & 0.7989 & 188.03 & 41.98 \\
\cmidrule(lr){2-9}
 & Qwen-4B-Instruction & 57.15 & 69.25 & 57.75 & 60.90 & 0.8094 & 218.89 & 37.25 \\
 & gemini-3-flash-preview & 85.38 & 98.77 & 92.42 & 91.31 & 0.8400 & 185.77 & 47.26 \\
\cmidrule(lr){2-9}
 & Geocode (v3.3\_pro) & \cellcolor{red!20}98.00 & \cellcolor{red!20}99.25 & \cellcolor{red!20}98.35 & \cellcolor{red!20}98.25 & 0.8429 & \cellcolor{red!20}127.13 & \cellcolor{red!20}19.36 \\
 & GeocodeLLM & 71.18 & 80.00 & 68.18 & 73.36 & \cellcolor{red!20}0.8757 & 144.10 & 22.34 \\
\bottomrule
\multicolumn{9}{l}{\footnotesize SC: Structural Consistency, PP: Points Presence, SCp: Segments Completeness, LO: LLM Overall Score (maximum score is 100).} \\
\multicolumn{9}{l}{\footnotesize SSIM: Structural Similarity Index Measure (maximum score is 1).} \\
\multicolumn{9}{l}{\footnotesize HD: Hausdorff Distance, CD: Chamfer Distance (lower values closer to 0 indicate better performance).} \\
\end{tabular}
\end{table*}

Detailed comparison results are presented in the table \ref{tab:consistency_results_highlighted}.

\textbf{Overall Performance:} In terms of visual metrics, \textit{Geo-code} achieves an average score exceeding 96 across different datasets, indicating its ability to precisely restore the geometric topology of the original images. Regarding pixel-level metrics, the SSIM exceeds 0.83, demonstrating high stylistic consistency with the ground truth. The CD scores range from 6.41 to 19.36, implying that the average pixel deviation is strictly controlled. Notably, the maximum HD of 127 pixels indicates that even in the worst-case scenarios, our method avoids catastrophic structural failures or significant geometric hallucinations.

\textbf{Dataset-Specific Analysis:}
\begin{itemize}
    \item On \textbf{AuxSolidMath}, \textit{Geo-code} achieves its best performance (e.g., LO of 97.84 and CD of 6.41), proving its proficiency in identifying complex 3D structures.
    \item On \textbf{GeoSketch} and \textbf{GeoQA}, the LO scores (97.85 and 98.20) and CD scores (12.04 and 14.41) confirm that the framework can accurately reconstruct planar geometries and their corresponding specialized markers.
    \item On \textbf{MathVerse}, the LO of 98.25 and CD of 19.36 demonstrate strong generalization capabilities when dealing with non-geometric elements like function curves.
\end{itemize}

\textbf{Comparison with Baselines:} Interestingly, while \textit{MatPlotCode} slightly exceeds \textit{Geo-code} in SC and SCp on the \textit{AuxSolidMath} dataset (+1.49 and +0.99 respectively), its pixel-level performance (CD and HD) is significantly inferior (-40.85 and -125.21). This discrepancy suggests that while baseline models may produce images that \textit{look} visually similar to the original, they often fail to capture the underlying geometric rigor and structural accuracy. The downstream reasoning benefits of our more precise reconstruction are further discussed in Section~\ref{sec:reasoning_exp}.

\textbf{Effectiveness of Geo-codeLM:} The fine-tuned \textit{Geo-codeLM} demonstrates superior performance, with the LO on \textit{AuxSolidMath} improving by +34.13\% over the baseline. Significant gains in CD and HD (e.g., CD on \textit{AuxSolidMath} dropping from 66.89 to 15.13) indicate that the model successfully learned to identify critical geometric skeletons. Furthermore, \textit{Geo-codeLM} outperforms \textit{FigCodifier} and \textit{MatPlotCode}—models fine-tuned on over 1M samples—despite using a smaller dataset. This highlights that \textbf{data quality and geometric logic are more critical than sheer volume} in image reconstruction tasks, validating the effectiveness of our synthesized high-quality training data.

\end{document}